\documentclass{article}

\usepackage{arxiv}
\newif\ifarxiv\arxivtrue

\usepackage[section]{placeins}
\usepackage{float}
\usepackage[utf8]{inputenc}
\usepackage[T1]{fontenc}
\usepackage{charter}
\usepackage[light]{roboto}
\usepackage{microtype}
\RequirePackage[x11names]{xcolor}
\usepackage{colortbl}

\usepackage{amsmath}
\usepackage{amssymb}
\usepackage{graphicx}
\usepackage{url}
\usepackage{hyperref}
\usepackage{nameref}
\hypersetup{% setup the hyperref-package options
    plainpages=false,           %   -
    colorlinks=false,           %   - colorize links?
    pdfborder={0 0 0},          %   -
    breaklinks=true,            %   - allow line break inside links
    bookmarksnumbered=true,     %
    bookmarksopen=true,         %
	hypertexnames=true,         %
}

\usepackage{multicol}
\usepackage{multirow}
\usepackage{pbox}
\usepackage{longtable}
\usepackage{booktabs}
\usepackage{tabu}
\usepackage{csvsimple}
\usepackage{array}
\usepackage[usestackEOL]{stackengine}
\usepackage[binary-units=true]{siunitx}
\usepackage{caption}
\usepackage{subcaption}
\usepackage{float}
\usepackage{paralist}
\usepackage{nicefrac}
\usepackage{algorithm}
\usepackage[noend]{algpseudocode}

\definecolor{t_gray}{HTML}{888888}
\definecolor{t_blue}{HTML}{355fb3}
\definecolor{t_red}{HTML}{b33535}
\definecolor{t_green}{HTML}{3bb335}
\definecolor{t_yellow}{HTML}{b39735}
\definecolor{t_darkgray}{HTML}{454545}
\definecolor{t_darkblue}{HTML}{1e3666}
\definecolor{t_darkgreen}{HTML}{22661e}
\definecolor{t_darkred}{HTML}{661e1e}
\definecolor{t_darkyellow}{HTML}{66571e}
\definecolor{t_lightblue}{HTML}{8ea7d7}
\definecolor{t_lightred}{HTML}{dc8989}
\definecolor{t_lightgreen}{HTML}{8ddc89}

\usepackage{pgfplots}
\usepackage{pgfplotstable}
\pgfplotsset{compat=1.14}
\usepgfplotslibrary{dateplot, statistics}
\pgfplotsset{
    cycle list={t_blue\\t_red\\t_green\\t_yellow\\},
}

\usepackage{enumitem}
\usepackage{mathtools}
\usepackage{amsthm}
\usepackage{thmtools}
\usepackage{bm}
\usepackage{bbm}
\usepackage{stmaryrd}

\usepackage[capitalise,nameinlink]{cleveref}

\usepackage{acro}
\acsetup{format/first-long=\slshape} % chktex 6
\acsetup{barriers/use=true}
\acsetup{barriers/reset=true}
\acsetup{single}

\preto\tabular{\setcounter{rownumbercount}{0}}
\newcounter{rownumbercount}

\usepackage[sort&compress,numbers]{natbib}

\DeclareMathOperator{\sign}{sign}
\newcommand*{\condbold}[3][]{\ifthenelse{\equal{#2}{1}#1}{\mathbf{#3}}{#3}}
\newcommand*{\evalres}[4][]{%
	\ifthenelse{\equal{#3}{m}}{\acs*{oom}}{%
	\ifthenelse{\equal{#3}{t}}{\acs*{oot}}{$\condbold[#1]{#2}{#3} \pm #4$}}%
}
\newcommand*\circled[2][1pt]{\tikz[baseline=(char.base)]{ % chktex 36
    \node[shape=circle,draw,inner sep=#1] (char) {#2};}}
\def\gcircled#1\gcircled{\circled{\small#1}}

\def\paperTitle{Ranking Structured Objects with \mbox{Graph Neural Networks}}
\title{\paperTitle}
\date{}

\author{
  Clemens Damke \\
  Heinz Nixdorf Institute \\
  Paderborn University \\
  \href{mailto:cdamke@mail.upb.de}{\texttt{cdamke@mail.upb.de}}
  \And
  Eyke Hüllermeier \\
  Institute of Informatics \\
University of Munich \\
  \href{mailto:eyke@ifi.lmu.de}{\texttt{eyke@ifi.lmu.de}}
}

\newcommand{\dac}[3]{\DeclareAcronym{#1}{short = #2, long = #3}}

\dac{nn}{NN}{neural network}
\dac{lm}{LRM}{logistic regression model}
\dac{svm}{SVM}{support vector machine}
\dac{mlp}{MLP}{multilayer perceptron}
\dac{cnn}{CNN}{convolutional neural network}
\dac{ltr}{LtR}{Learning to Rank}

\dac{gcr}{GC/GR}{graph classification and regression}
\dac{gc}{GC}{graph classification}
\dac{gr}{GR}{graph regression}
\dac{gk}{GK}{graph kernel}
\dac{gnn}{GNN}{graph neural network}
\dac{gcnn}{GCNN}{graph convolutional neural network}
\dac{gcn}{GCN}{graph convolutional network}
\dac{sage}{GraphSAGE}{\textsc{sa}mple and aggre\textsc{g}at\textsc{e}}
\dac{gin}{GIN}{graph isomorphism network}
\dac{wl2gnn}{2-WL-GNN}{2-dimensional Weisfeiler-Lehman GNN}
\dac{sagpool}{SAGPooling}{self-attention graph pooling}
\dac{sampool}{SAMPooling}{softmax attention mean pooling}
\dac{wlst}{WL\textsubscript{ST}}{Weisfeiler-Lehman subtree kernel}
\dac{wlsp}{WL\textsubscript{SP}}{Weisfeiler-Lehman shortest-path kernel}

\dac{cmpnn}{CmpNN}{Comparative Neural Network}

\dac{ml}{ML}{machine learning}
\dac{xai}{XAI}{explainable artificial intelligence}
\dac{nlp}{NLP}{natural language processing}
\dac{gi}{GI}{graph isomorphism}
\dac{wl}{WL}{Weisfeiler-Lehman}
\dac{ft}{FT}{Fourier transform}
\dac{bfs}{BFS}{breadth-first-search}
\dac{lcm}{LCM}{lowest common multiple}
\dac{dp}{DP}{discriminative power}
\dac{cp}{CP}{computational power}
\dac{ndcg}{NDCG}{normalized discounted cummulative gain}
\dac{mse}{MSE}{mean squared error}

\dac{gpu}{GPU}{graphics processing unit}
\dac{nci}{NCI}{National Cancer Institute}
\dac{sse}{SSE}{secondary structure element}
\dac{oom}{OOM}{out of memory}
\dac{oot}{OOT}{out of time}
\dac{cfg}{CFG}{control-flow graph}
\dac{ogb}{OGB}{Open Graph Benchmark}

\begin{document}
\maketitle
%{\vskip -0.2in\centering Heinz Nixdorf Institute and Department of Computer Science\\Paderborn University\vskip 0.2in}

%!TEX root = ../arxiv_paper.tex
\begin{abstract}
\Acp{gnn} have been successfully applied in many structured data domains, with applications ranging from molecular property prediction to the analysis of social networks.
Motivated by the broad applicability of \acp{gnn}, we propose the family of so-called \emph{RankGNNs}, a combination of neural \ac{ltr} methods and \acp{gnn}.
RankGNNs are trained with a set of pair-wise preferences between graphs, suggesting that one of them is preferred over the other.
One practical application of this problem is drug screening, where an expert wants to find the most promising molecules in a large collection of drug candidates.
We empirically demonstrate that our proposed pair-wise RankGNN approach either significantly outperforms or at least matches the ranking performance of the na\"ive point-wise baseline approach, in which the \ac{ltr} problem is solved via \ac{gnn}-based graph regression.
\keywords{Graph-structured data \and Graph neural networks \and  Preference learning \and Learning to rank}
\end{abstract}

%!TEX root = ../arxiv_paper.tex
% chktex-file 46
\section{Introduction}%
\label{sec:intro}

Bringing a set of objects $o_1, \dots, o_N$ into a particular order is an important problem with many applications, ranging from task planning to recommender systems.
In such domains, the criterion defining the underlying order relation $\succeq$ typically depends on properties (features) of the objects (for example the price and quality of a product). If the sorting criterion (and hence the relation $\succeq$) is not explicitly given, one may think of inferring it from exemplary data, often provided in the form of a set of pair-wise orderings $o_i \succeq o_j$ (e.g., representing that the user prefers product $o_i$ over product $o_j$). This gives rise to a machine learning task often referred to as \ac{ltr}. Thus, the goal is to learn a general ordering strategy (preference model) from sample data of the above kind, which can then be used to sort any new (previously unseen) set of objects.
%This is called the \ac{ltr} problem.

While existing state-of-the-art \ac{ltr} approaches assume that objects $o_i$ are represented by feature vectors $x_i \in \mathbb{R}^n$, in this paper, we will consider the \ac{ltr} problem for another quite natural and practically important representation, namely the domain of finite graphs. Methods for learning to rank objects represented in the form of graphs can, for example, be used in applications such as drug screening, where the ranked objects are the molecular structures of drug candidates.

To support the ranking of structured objects such as graphs, existing \ac{ltr} methods need to be adapted. Previously, \citet{Agarwal2010} has considered the problem of ranking the vertices within a given graph.
However, to the best of our knowledge, the graph-wise \ac{ltr} problem has so far only been described in the context of specific domains, such as drug discovery, where manually chosen graph feature representations were used~\citep{Zhang2015}.
Motivated by the success of \acp{gnn} in graph representation learning, we propose a simple architecture that combines \acp{gnn} with neural \ac{ltr} approaches.
The proposed approach allows for training ranking functions in an end-to-end fashion and can be applied to arbitrary graphs without the need to manually choose a domain-specific graph feature representation.

Our neural graph ranking architecture will be introduced in \cref{sec:rankgnn}. Before, 
the \ac{ltr} and \ac{gnn} models that are used in this architecture are described in \cref{sec:ranking} and \cref{sec:gnn}, respectively.
In \cref{sec:eval}, we evaluate our approach on a selection of graph benchmark datasets.

\section{Object Ranking}%
\label{sec:ranking}

\ac{ltr} approaches are often categorized as point-wise, pair-wise, and list-wise methods. 
%There are three main families of \ac{ltr} approaches for objects:
%\begin{enumerate*}
%    \item Point-wise,
%    \item pair-wise and
%    \item list-wise.
%\end{enumerate*}
We begin with a short overview of these families.
Afterwards, a more in-depth introduction is given to a selection of neural pair-wise approaches that we shall built upon in \cref{sec:rankgnn}.

\subsection{Overview of LtR Approaches}%
\label{sec:ranking:overview}

\paragraph{Point-wise methods} assume the existence of a (latent) utility function representing the sought preference relation $\succeq$, i.e., 
that an ordinal or numeric utility score $u_i \in \mathbb{R}$ can be assigned to each object $o_i \in O$ such that 
$$
\forall \, o_i, o_j \in O : \, {u_i \geq u_j} \Leftrightarrow {o_i \succeq o_j} \, .
$$ 
Based on training data in the form of exemplary (and possibly noisy) ratings, i.e., object/utility pairs $\{ (x_i , u_i) \}_{i=1}^N \subset X \times \mathbb{R}$, where $x_i \in X$ is the feature representation of $o_i$, the \ac{ltr} problem can be solved by fitting a model $f_u: X \to \mathbb{R}$ using standard ordinal or numeric regression methods. Given a new set of objects $\{ o_j'\}_{j=1}^M$ to be ranked, these objects are then sorted in decreasing order of their estimated utilities $f_u(o_j')$. 
Note that point-wise methods are restricted to linear orders but cannot represent more general relations, such as partial orders.
%However, as alluded to in the introduction, there are situations where object utilities are unknown. Moreover, in cases where the $\succeq$ relation is a partial ordering or possibly even intransitive, object utilities are not only unknown, but fundamentally unknowable, since utility scores can only describe total ordering relationships.

\paragraph{Pair-wise methods} proceed from training data in the form of a set of ordered object pairs $S = {\{ o_{a_i} \succeq o_{b_i} \}}_{i=1}^{N}$, i.e., \emph{relative} training information in the form of pair-wise comparisons rather than absolute assessments.
Based on such training samples $S$, the goal is to learn the underlying preference relation $\succeq$.
The resulting model $f_{\succeq}: O \times O \to \{0, 1\}$ is a binary classifier, which is supposed to return $f_{\succeq}(o_i, o_j) = 1$ iff $o_i \succeq o_j$.

One of the first pair-wise preference methods was the Ranking~SVM~\citep{Joachims2002}\,---\,essentially a standard \ac{svm} trained on the differences between vector representations of object preference pairs.
Later, \citet{Burges2005} proposed the RankNet architecture, which is also trained using feature vector differences but uses a \ac{mlp} instead of an \ac{svm}.
Since then, multiple extensions of those approaches have been developed~\citep{Burges2010}.
One commonality between all of them is their training optimization target, namely to minimize the number of predicted inversions, i.e., the number of pairs $o_i \succeq o_j$ with $f_{\succeq}(o_i, o_j) = 0$.
An important difference between existing pair-wise approaches concerns the properties they guarantee for the learned preference relation; three properties commonly considered are 
\begin{itemize}
\item reflexivity ($\forall x:\, x \succeq x$), 
\item antisymmetry ($\forall x, y:\, x \nsucceq y \Rightarrow y \succeq x$), and 
\item transitivity ($\forall x, y, z:\, (x \succeq y \land y \succeq z) \Rightarrow x \succeq z$).
\end{itemize}
The set of desirable properties depends on the domain. While some approaches guarantee that the learned relation fulfills all three properties~\citep{Koeppel2019}, others, for example, explicitly allow for non-transitivity~\citep{Rigutini2011}.

Assuming a suitable pair-wise ranking model $f_{\succeq}$ was selected and trained, one then typically wants to produce a ranking for some set of objects $\{o_i' \}_{i=1}^M$. To this end, a ranking (rank aggregation) procedure is applied to the preferences predicted for all pairs $(o_i',o_j')$. A simple example of such a procedure is to sort objects $o_i$ by their Borda count 
$c_i = \sum_{j \neq i} f_{\succeq}(o_i, o_j)$,
i.e., by counting how often each object $o_i$ is preferred over another object.  
%a ranking can then be obtained by sorting the objects according to their preference counts.
Alternatively, the classifier $f_{\succeq}$ can also be used directly as the comparator function in a sorting algorithm;
this reduces the number of comparisons from $\mathcal{O}(M^2)$ to $\mathcal{O}(M \log M)$.
While the latter approach is much more efficient, it implicitly assumes that $f_{\succeq}$ is transitive.
The rankings produced by an intransitive sorting comparator are generally unstable, because they depend on the order in which the sorting algorithm compares the objects \cite{mpub383}.
This might not be desirable in some domains.

\paragraph{List-wise methods} generalize the pair-wise setting.
Instead of determining the ordering of object pairs, they directly operate on complete rankings (lists) of objects, training a model based on a list-wise ranking loss function.
One of the first list-wise losses was proposed by \citet{Cao2007}. Given a set $S$ of objects, their ListNet approach uses a probability distribution over all possible rankings of $S$ and is trained by minimizing the cross-entropy between the model's current ranking distribution and some target distribution.
Compared to pair-wise approaches, list-wise methods exhibit a higher expressivity, which can be useful to capture effects such as context-dependence of preferences \citep{Pfannschmidt2018}.
%proposed list-wise architectures that are able to capture context-dependent preferences. Given two sets of objects $S = \{o_1, o_2, o_3\}$ and $S' = \{o_1, o_2, o_4\}$, a context-dependent preference could for example produce the ordering $o_2 \succeq o_1 \succeq o_3$ for $S$ and the ordering $o_1 \succeq o_2 \succeq o_4$ for $S'$, i.e., the preference between $o_1$  and $o_2$ changes depending on whether they are compared in the presence of $o_3$ or in the presence of $o_4$. One real-world domain in which such context dependencies can be observed is psychology: Whether a customer prefers product A over product B often depends on the set of other products to which A and B can be compared\,---\,this is also known as the \emph{decoy effect}~\citep{Huber1982}. Since context-dependent preferences cannot be captured by a pair-wise approach, a list-wise \ac{ltr} method can be a better choice in such domains.
In general, however, if this level of expressiveness is not required, recent results by \citet{Koeppel2019} suggest that the list-wise approaches have no general advantage over the (typically simpler) pair-wise methods.
To tackle the graph \ac{ltr} problem in \cref{sec:rankgnn}, we will therefore focus on the pair-wise approach.

\subsection{Neural Pair-wise Ranking Models}%
\label{sec:ranking:neural}

As already stated, we propose a combination of existing \ac{ltr} methods and \acp{gnn} to solve graph ranking problems.
Due to the large number of existing \ac{ltr} approaches, we will however not evaluate all possible combinations with \acp{gnn}, but instead focus on the following two representatives:
\begin{enumerate}[label=\textbf{\arabic*.}]
    \item \textbf{DirectRanker~\citep{Koeppel2019}:}
        A recently proposed generalization of the already mentioned pair-wise RankNet architecture~\citep{Burges2005}.
        It guarantees the reflexivity, antisymmetry, and transitivity of the learned preference relation and achieves state-of-the-art performance on multiple common \ac{ltr} benchmarks.
    \item \textbf{\acs*{cmpnn}~\citep{Rigutini2011}:}
        Unlike DirectRanker, this pair-wise architecture does not enforce transitivity.
        The authors suggest that this can, for example, be useful to model certain non-transitive voting criteria.
\end{enumerate}
%We will now take a look at how those two approaches work.
Formally, the DirectRanker architecture is defined as
\begin{equation}
    f^{\text{DR}}_{\succeq}(o_i, o_j) \coloneqq \sigma \left(w^{\top} (h(x_i) - h(x_j)) \right) \text{,}\label{eq:ranking:direct}
\end{equation}
where $x_i, x_j \in \mathbb{R}^{n}$ are feature vectors representing the compared objects $o_i, o_j$, the function $h: \mathbb{R}^{n} \to \mathbb{R}^{d}$ being a standard \ac{mlp}, $w \in \mathbb{R}^{d}$ a learned weight vector and an activation function $\sigma: \mathbb{R} \to \mathbb{R}$ such that
$\sigma(-x) = -\sigma(x)$ and $\sign(x) = \sign(\sigma(x))$ 
for all $x \in \mathbb{R}$.
One could, for example, use $\sigma = \tanh$ and interpret negative outputs of $f^{\text{DR}}_{\succeq}(o_i, o_j)$ as $o_j \succeq o_i$ and positive outputs as $o_i \succeq o_j$.
This model can be trained in an end-to-end fashion using gradient descent with the standard binary cross-entropy loss.
Note that $f^{\text{DR}}_{\succeq}$ can be rewritten as $\sigma(f^{\text{DR}}_u(x_i) - f^{\text{DR}}_u(x_j))$, with $f^{\text{DR}}_u(x) \coloneqq w^{\top} h(x)$.
DirectRanker therefore effectively learns an object utility function $f^{\text{DR}}_u$ and predicts $o_i \succeq o_j$ iff $f^{\text{DR}}_u(x_i) \geq f^{\text{DR}}_u(x_j)$.
Thus, the learned preference relation $f^{\text{DR}}_{\succeq}$ directly inherits the reflexivity, antisymmetry and transitivity of the $\geq$ relation.
The main difference between DirectRanker and a point-wise regression model is that DirectRanker learns $f^{\text{DR}}_u$ indirectly from a set of object preference pairs.
Consequently, DirectRanker is not penalized if it learns some order-preserving transformation of $f^{\text{DR}}_u$.
We will come back to this point in \cref{sec:eval:utilities}.

Let us now look at the so-called \ac{cmpnn} architecture, which generalizes the DirectRanker approach.
The main difference between both is that \ac{cmpnn} does not implicitly assign a score $f_u(x_i)$ to each object $o_i$.
This allows it to learn non-transitive preferences.
\acp{cmpnn} are defined as follows:
\begin{gather}
    f^{\text{Cmp}}_{\succeq}(o_i, o_j) \coloneqq \sigma(z_{\succeq} - z_{\preceq})\textrm{, with} \label{eq:ranking:cmp}\\
    \begin{aligned}
        z_{\succeq} &\coloneqq \tau(w_{1}^{\top} z_1 + w_{2}^{\top} z_2 + b'), &
        z_1 &\coloneqq \tau(W_{1} x_i + W_{2} x_j + b),\\
        z_{\preceq} &\coloneqq \tau(w_{2}^{\top} z_1 + w_{1}^{\top} z_2 + b'), &
        z_2 &\coloneqq \tau(W_{2} x_i + W_{1} x_j + b) \text{.}
    \end{aligned}\nonumber
\end{gather}
Here, $w_1, w_2 \in \mathbb{R}^d$ and $W_1, W_2 \in \mathbb{R}^{d \times n}$ are shared weight matrices, $b, b'$ bias terms, and $\sigma, \tau$ activation functions.
Intuitively, $z_{\succeq} \in \mathbb{R}$ and $z_{\preceq} \in \mathbb{R}$ can be interpreted as weighted votes towards the predictions $o_i \succeq o_j$ and $o_j \succeq o_i$, respectively.
A \ac{cmpnn} will simply choose the alternative with the largest weight.
The key idea behind the definitions in (\ref{eq:ranking:cmp}) is that the pairs $z_{\succeq}, z_{\preceq}$ and $z_1, z_2$ will swap values when swapping the compared objects $o_i, o_j$.
Consequently, $f^{\text{Cmp}}_{\succeq}$ must be reflexive and antisymmetric~\citep[see][]{Rigutini2011}.
If we set $W_1 = w_1 = 0$, the voting weights $z_{\succeq}, z_{\preceq} \in \mathbb{R}$ reduce to the predictions of a standard \ac{mlp} $h$ with the input $o_i$ and $o_j$, respectively, i.e., $z_{\succeq} = h(x_i)$ and $z_{\preceq} = h(x_j)$.
In this case, the \ac{cmpnn} effectively becomes a DirectRanker model.
By choosing non-zero weights for $W_1$ and $w_1$, the model can however also learn non-transitive dependencies between objects.
In fact, \citeauthor{Rigutini2011} have shown that \acp{cmpnn} are able to approximate almost all useful pair-wise preference relations~\citep[Thm.~1]{Rigutini2011}.

\section{Graph Neural Networks}%
\label{sec:gnn}

Over the recent years, \acp{gnn} have been successfully employed for a variety of graph ML tasks, with applications ranging from graph classification and regression to edge prediction and graph synthesis.
Early \ac{gnn} architectures were motivated by spectral graph theory and the idea of learning eigenvalue filters of graph Laplacians~\citep{Bruna2013,Henaff2015}.
Those spectral \acp{gnn} take a graph $G = (V, E)$ with vertex feature vectors $x_i \in \mathbb{R}^n$ as input and iteratively transform those vertex features by applying a filtered version of the Laplacian $L$ of $G$.
Formally, the filtered Laplacian is defined as $\hat{L} = U^{\top} g(\Lambda) U$, where $L = {U^{\top} \Lambda U}$ is an eigendecomposition of $L$ and $g$ is a learned eigenvalue filter function that can amplify or attenuate the eigenvectors $U$.
Intuitively, spectral \acp{gnn} learn which structural features of a graph are important and iteratively aggregate the feature vectors of the vertices that are part of a common important structural graph feature.
Each of those aggregations is mathematically equivalent to a convolution operation. This is why they are referred to as \emph{(graph) convolution layers}.

One important disadvantage of spectral convolutions is their computational complexity, making them especially unsuitable for large graphs.
To overcome this limitation, \citeauthor{Kipf2017} proposed the so-called \ac{gcn} architecture~\citep{Kipf2017}, which restricts the eigenvalue filters $g$ to be linear.
As a consequence of this simplification, only adjacent vertices need to be aggregated in each convolution.
Formally, the simplified \ac{gcn} convolution can be expressed as follows:
\begin{equation}
    x'_i = \sigma\left(W \left( \eta_{ii}\, x_i + \smashoperator[lr]{\sum_{v_j \in \Gamma(v_i)}} \eta_{ij}\, x_j \right)\right) \label{eq:gnn:gcn}
\end{equation}
Here, $x_i, x'_i \in \mathbb{R}^d$ are the feature vectors of $v_i \in V$ before and after applying the convolution, $\Gamma(v_i)$ is the set of neighbors of $v_i$, $W \in \mathbb{R}^{d \times n}$ is a learned linear operator representing the filter $g$, $\sigma$ some activation function, and $\eta_{ii}, \eta_{ij} \in {[0,1]}$ normalization terms that will not be discussed here.
After applying a series of such convolutions to the vertices of a graph, the resulting convolved vertex features can be used directly to solve vertex-level prediction tasks, e.g.\ vertex classification.
To solve graph-level problems, such as graph classification or graph ranking, the vertex features must be combined into a single graph vector representation.
This is typically achieved via a pooling layer, which could, for example, simply compute the component-wise mean or sum of all vertex features.
More advanced graph pooling approaches use sorting or attention mechanisms in order to focus on the most informative vertices~\citep{Zhang2018,Lee2019}.

\Citet{Xu2018} show that restricting the spectral filter $g$ to be linear not only reduces the computational complexity but also the discriminative power of the \ac{gcn} architecture.
More precisely, they prove that \text{any} \ac{gnn} using a vertex neighborhood aggregation scheme such as (\ref{eq:gnn:gcn}) can \emph{at most} distinguish those graphs that are distinguishable via the so-called 1-dimensional \ac{wl} graph isomorphism test~\citep{Cai1992}.
\Acp{gcn} do, in fact, have a strictly lower discriminative power than 1-\acs{wl}, i.e., there are 1-\acs{wl} distinguishable graphs, which will always be mapped to the same graph feature vector by a \ac{gcn} model.
In addition to this bound, \citet{Xu2018} also propose the \ac{gin} architecture, which is able to distinguish \emph{all} 1-\acs{wl} distinguishable graphs.
% \begin{equation}
%     x'_i = g\left((1 + \varepsilon) x_i + \smashoperator[lr]{\sum_{v_j \in \Gamma(v_i)}} x_j \right)
% \end{equation}
% Here, $g: \mathbb{R}^n \to \mathbb{R}^d$ is a standard \ac{mlp} and $\varepsilon > 0$ some small irrational constant, which is formally required but typically set to $0$ in practice.
Recently, multiple approaches going beyond the 1-\acs{wl} bound have been proposed.
The so-called \acs{wl2gnn} architecture, for example, is directly based on the 2-dimensional (Folklore) \ac{wl} test~\citep{Damke2020}.
Other current approaches use higher-order substructure counts~\citep{Bouritsas2020} or so-called $k$-order invariant networks~\citep{Maron2019}.

\section{Neural Graph Ranking}%
\label{sec:rankgnn}

To tackle the graph \ac{ltr} problem, we propose the family of \emph{RankGNN} models.
A RankGNN is a combination of a \ac{gnn} and one of the existing neural \ac{ltr} methods.
The \ac{gnn} component is used to embed graphs into a feature space.
The embedded graphs can then be used directly as the input for a comparator network, such as DirectRanker~\citep{Koeppel2019} or \ac{cmpnn}~\citep{Rigutini2011}.
Formally, a RankGNN is obtained by simply using a \ac{gnn} to produce the feature vectors $x_i, x_j$ in (\ref{eq:ranking:direct}) and (\ref{eq:ranking:cmp}) for a given pair of graphs $G_i, G_j$.
Since all components of such a combined model are differentiable, the proposed RankGNN architecture can be trained in an end-to-end fashion.
Despite the simplicity of this approach, there are a few details to consider when implementing it;
these will be discussed in the following sections.

\subsection{Efficient Batching for RankGNNs}%
\label{sec:rankgnn:batch}
\begin{figure*}[t]
    \centering
    \includegraphics[width={\ifarxiv0.85\linewidth\else0.85\linewidth\fi}]{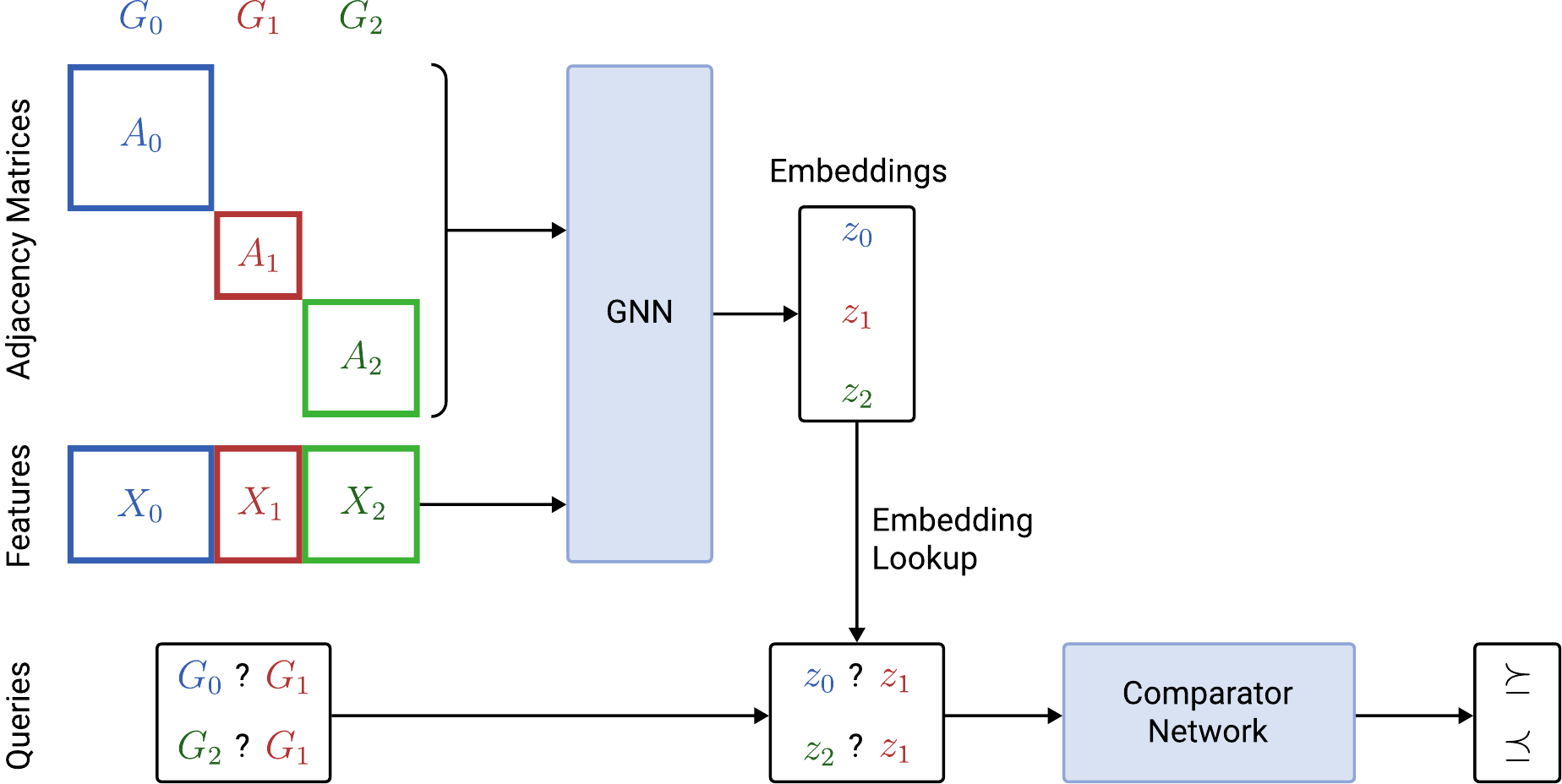}
    \caption{
        General architecture of the proposed family of RankGNNs.
        Here the common sparse adjacency representation for message-passing \acp{gnn} is shown; different types of graph batch encodings can of course also be used.
    }\label{fig:rankgnn-architecture}
\end{figure*}
In the existing neural \ac{ltr} approaches for objects $o_i$ that are represented by features $x_i \in \mathbb{R}^n$, efficient batch training is possible by encoding a batch of $k$ relations ${\{ o_{a_i} \succeq o_{b_i} \}}_{i=1}^{k}$ with two matrices 
$$
A \coloneqq \begin{psmallmatrix}x_{a_1}\\\vphantom{\int\limits^x}\smash{\vdots} \\ x_{a_k}\end{psmallmatrix} \in \mathbb{R}^{k \times n}, \quad B \coloneqq \begin{psmallmatrix}x_{b_1}\\\vphantom{\int\limits^x}\smash{\vdots} \\ x_{b_k}\end{psmallmatrix} \in \mathbb{R}^{k \times n}
$$ 
and using 
$$
Y \coloneqq \begin{psmallmatrix}1\\\vphantom{\int\limits^x}\smash{\vdots} \\ 1\end{psmallmatrix} \in \mathbb{R}^{k}
$$ 
as the target prediction of the model.
However, this approach is suboptimal in the graph \ac{ltr} setting.
Given the relations $\{G_0 \succeq G_1, G_1 \succeq G_2\}$, the graph $G_1$ would for example have to be encoded twice.
When dealing with datasets that consist of possibly large graphs, such redundant encodings quickly become infeasible due to the additional memory and runtime requirements incurred by the \ac{gnn} having to embed the same graph multiple times.
To prevent this redundancy, each graph occurring on the left or the right side of a relation should instead only be encoded once as part of a single graph batch.
This graph batch can be fed directly into a \ac{gnn} to produce a matrix $Z$ of graph feature embeddings.
The individual graph relation pairs $G_i \succeq G_j$ can then be simply represented as pairs of indices $(i, j)$ pointing to the corresponding feature vectors in the embedding matrix $Z$.
Using those pointers, the graph vector representations for each pair can be looked up in $Z$.
\Cref{fig:rankgnn-architecture} illustrates this idea.

\subsection{Sorting Graphs with RankGNNs}%
\label{sec:rankgnn:sort}
After training a RankGNN model using a set of graph relation pairs, the model can be used to compare arbitrary graph pairs.
Following the approach of \citet{Koeppel2019} and \citet{Rigutini2011}, a set of graphs can then be ordered by using the RankGNN as the comparator function in a standard sorting algorithm.
We propose a simple parallelized quicksort-based scheme to sort graphs.
When implementing a RankGNN model on a parallel compute device, such as a \acs{gpu}, there is a constant time overhead for each call to the model.
To reduce the total cost of this call overhead, we suggest that all pivot comparison queries in one layer of the recursive quicksort call tree should be evaluated by the RankGNN in parallel.
Using this parallel comparison approach, only one model invocation is required for each layer of the call tree, i.e., the asymptotic model call overhead for sorting $n$ graphs is in $O(\log n)$.
Additionally, a more efficient approach is available for DirectRanker-based models. 
There, the implicitly learned utility function $f^{\text{DR}}_u$ can be computed directly for a set of graphs.
A standard sorting algorithm can then be applied without any further calls to the model, which reduces the call overhead to $O(1)$.
% \begin{algorithm}
% 	\caption{Graph Sorting via RankGNNs}\label{algo:rankgnn:sort}
% 	\begin{algorithmic}[1]
%  		\Function{sort}{$G = {(G_1, \dots, G_n)}, h_{\succeq}: \mathcal{G}^2 \to {\{0,1\}}$}
%             \State{$\mathit{sorted} \leftarrow (\texttt{nil} \times n)$}
%             \Comment{Empty result array}
% 	        \State{$\mathit{parts} \leftarrow (G)$}
% 	        \Comment{Partitions}
% 	        \While{$\mathit{parts} \neq ()$}
% 	            \State{$\mathit{parts}' \leftarrow ()$}
% 	            \ForAll{$\mathit{part} \in \mathit{parts}$}
% 	                \State{$p, \mathit{rest} \leftarrow \mathit{part}[0], \mathit{part}[1:]$}
% 	            \EndFor{}
% 	        \EndWhile{}
% 	        \State{\Return{$\mathit{sorted}$}}
% 		\EndFunction{}
% 	\end{algorithmic}
% \end{algorithm}

\section{Evaluation}%
\label{sec:eval}

To evaluate the family of RankGNNs described in \cref{sec:rankgnn}, we choose six different combinations of \acp{gnn} and comparator networks.
The evaluated graph embedding modules are \acs{gcn}~\citep{Kipf2017}, \acs{gin}~\citep{Xu2018}, and \acs{wl2gnn}~\citep{Damke2020}.
Those three \ac{gnn} methods are combined with the previously described DirectRanker~\citep{Koeppel2019} and the \ac{cmpnn}~\citep{Rigutini2011} comparator.
Because there are currently no common graph ranking benchmark datasets, we instead convert a selection of graph regression benchmarks into ranking problems by interpreting the numeric regression targets as utility values, which are used to determine the target orderings.
The following five graph regression datasets are used:
\begin{enumerate}[label={\textbf{\arabic*.}}]
    \item \textbf{TRIANGLES:}
        This is a synthetic dataset that we created.
        It consists of 778 randomly sampled graphs, each of which contains 3 to 85 unlabeled vertices.
        The regression target is to learn how many triangles, i.e.\ 3-cliques, a given graph contains.
        The triangle counts in the sampled graphs vary between 0 and 9.
        The sampled graphs are partitioned into $80\%/10\%/10\%$ training/validation/test splits. 
    \item \textbf{OGB-molesol, -mollipo and -molfreesolv:}
        These three datasets are provided as part of the \ac{ogb} project~\citep{Hu2020}.
        They contain 1128, 4200, and 642 molecular structure graphs, respectively.
        The regression task is to predict the solubility of a molecule in different substances.
        We use the dataset splits that are provided by \ac{ogb}.
    \item \textbf{ZINC:}
        This dataset contains the molecular structures of 250k commercially available chemicals from the ZINC database~\citep{Sterling2015}.
        The regression task is to predict the so-called \emph{octanol-water partition coeffients}.
        We use the preprocessed and presplit graphs from the TUDataset collection~\citep{Morris2020}.
\end{enumerate}
To train the proposed pair-wise graph ranking network architecture, a subset of graph pairs from the training split is sampled uniformly at random.
The size of a training sample is $M = \alpha N$, where $N$ is the number of graphs in the training split of a dataset and $\alpha \in \mathbb{R}^+$ is a constant factor.
We use a sampling factor of $\alpha = 20$ for all datasets except ZINC, where we use $\alpha = 3$ due to the large number of graphs in the training split ($N_{\textrm{ZINC}} = 220011$, whereas e.g.\ $N_{\textrm{OGB-mollipo}} = 3360$).
This sampling strategy guarantees that each training graph occurs in at least one sampled pair with a probability of at least $1 - e^{-2\alpha}$;
thus, for both $\alpha = 20$ and even $\alpha = 3$, all graphs are considered with high probability ($> 99.75\%$).

In addition to the six pair-wise RankGNN model variants, we also evaluate the ranking performance of standard point-wise \ac{gnn} graph regression models, which are trained directly on graph utility values.
We use two different target graph utilities:
The original regression target $y_i \in \mathbb{R}$ for each training graph $G_i$, and the \emph{normalized graph rank} $\bar{r}_i \in {[0,1]}$, i.e.\ the normalized ordinal index of each training graph $G_i$ when sorted by $y_i$. %($r_i = 0$ iff $y_i = \min_{j} y_j$ and $r_i = 1$ iff $y_i = \max_{j} y_j$).

\subsection{Experimental Setup}%
\label{sec:eval:setup}

We evaluate the performance of the different RankGNN variants via Kendall's $\tau_B$ rank correlation coefficient.
Given two graph rankings $r_1: \mathcal{G} \to \mathbb{N}$, $r_2: \mathcal{G} \to \mathbb{N}$, this coefficient is defined as 
$$
\tau_B \coloneqq \frac{C - D}{\sqrt{(C+D+T_1)(C+D+T_2)}} \, , 
$$
where $C$ is the number of concordant pairs 
$$
\{ \{G_i, G_j\} \mid i \neq j \land r_1(G_i) < r_1(G_j) \land r_2(G_i) < r_2(G_j) \} \, , 
$$
$D$ is the number of discordant pairs 
$$
\{ \{G_i, G_j\} \mid i \neq j \land r_1(G_i) < r_1(G_j) \land r_2(G_i) > r_2(G_j) \} \, ,
$$ 
and $T_{1,2}$ are the numbers of tied graph pairs, which have the same rank in $r_1$ and $r_2$, respectively.
Kendall's $\tau_B$ rank coefficient ranges between $-1$ and $+1$, where $\tau_B = +1$ indicates that the two compared rankings are perfectly aligned, whereas $\tau_B = -1$ means that one rankings is the reversal of the other. 

Another commonly used metric in the \ac{ltr} literature is the \ac{ndcg}, which penalizes rank differences at the beginning of a ranking more than differences at the end.
This is motivated by the idea that typically only the top-$k$ items in a ranking are of interest.%, while rank differences between objects at the end of a ranking are less important.
We do not employ the \ac{ndcg} metric because this motivation does not hold for the used target graph rankings.
Since the target rankings are derived from regression targets, such as the water solubility of a molecule, both, the beginning and the end of a ranking are of interest and should therefore be weighted equally.

To train the evaluated point- and pair-wise models, we use the standard Adam optimizer~\citep{Kingma2015}.
The \ac{mse} loss is used for the point-wise regression models, while the pair-wise variants of those \acp{gnn} are optimized via binary cross-entropy.
All models were tuned via a simple hyperparameter grid search over the following configurations:
\begin{enumerate}[label=\textbf{\arabic*.}]
    \item \textbf{Layer widths:} $\{32, 64\}$.
        The width of both, the convolutional layers, as well as the fully-connected \ac{mlp} layers that are applied after graph pooling.
    \item \textbf{Number of graph convolutions:} $\{3, 5\}$. A fixed number of two hidden layers was used for the \ac{mlp} that is applied after the pooling layer.
    \item \textbf{Pooling layers:} $\{\text{mean}, \text{sum}, \text{softmax}\}$.
        Here, ``mean'' and ``sum'' refer to the standard arithmetic mean and sum operators, as described by \citet{Xu2018}, while ``softmax'' refers to the weighted mean operator described by \citet{Damke2020}.
    \item \textbf{Learning rates:} $\{\num{e-2}, \num{e-3}, \num{e-4}\}$.
\end{enumerate}
We used standard sigmoid activations for all models and trained each hyperparameter configuration for up to 2000 epochs with early stopping if the validation loss did not improve by at least \num{e-4} for 100 epochs.
The configuration with the highest $\tau_B$ coefficient on the validation split was chosen for each model/dataset pair.
To account for differences caused by random weight initialization, the training was repeated three times;
10 repeats were used for the TRIANGLES dataset due to its small size and fast training times.
Note that, depending on the type of \ac{gnn}, the pair-wise models can have between $3\%$ and $10\%$ more trainable weights than their point-wise counterparts, due to the added comparator network.
All models were implemented in Tensorflow and trained using a single Nvidia GTX~1080Ti \acs{gpu}.
The code is available on GitHub\footnote{%
\url{https://github.com/Cortys/rankgnn}%
% \url{https://github.com/...} (anonymized)%
}.

\subsection{Discussion of Results}%
\label{sec:eval:results}

\cref{tbl:eval:results} shows the ranking performance of the evaluated point- and pair-wise approaches on the test splits of the previously described benchmark datasets.
Each group of rows corresponds to one of the three evaluated \ac{gnn} variants.
The first two rows in each group show the results for the point-wise models that are trained directly on the original regression targets and on the normalized ranks, respectively.
The last two rows in each group hold the results for the pair-wise DirectRanker- and \acs{cmpnn}-based models.
%The pair-wise \acp{gnn} with the \textit{Dr} prefix use a DirectRanker model as their comparator network, the ones with the \textit{Cmp} prefix use a \ac{cmpnn}.
Generally speaking, the pair-wise approaches either significantly outperform or at least match the performance of the point-wise regression models.
The most significant performance delta between the point- and pair-wise approaches can be observed on the ZINC and OGB-mollipo datasets.
Only on the OGB-molesol dataset, the point-wise models achieve a slightly higher average $\tau_B$ value than the pair-wise models, which is however not significant when considering the standard deviations.
Overall, we find that the pair-wise rank loss that directly penalizes inversions is much better suited for the evaluated graph ranking problems than the point-wise \ac{mse} loss.

Comparing the two evaluated variants of point-wise regression models, we find that the ones trained on normalized graph ranks generally either have a similar or significantly better ranking performance than the regression models with the original targets.
We will come back to this difference in \cref{sec:eval:utilities}.
\begin{table*}[t]
	\caption{
        Mean Kendall's $\tau_B$ coefficients with standard deviations for the rankings produced by point- and pair-wise models on unseen test graphs.
    }\label{tbl:eval:results}
	\centering
	{\ifarxiv\footnotesize\else\scriptsize\fi\begin{filecontents*}{data/results.csv}
id;model;params;isDefault;triangleTestMean;triangleTestStd;triangleTrainMean;triangleTrainStd;triangleBestTest;triangleBestTrain;molesolTestMean;molesolTestStd;molesolTrainMean;molesolTrainStd;molesolBestTest;molesolBestTrain;mollipoTestMean;mollipoTestStd;mollipoTrainMean;mollipoTrainStd;mollipoBestTest;mollipoBestTrain;molfreesolvTestMean;molfreesolvTestStd;molfreesolvTrainMean;molfreesolvTrainStd;molfreesolvBestTest;molfreesolvBestTrain;zincTestMean;zincTestStd;zincTrainMean;zincTrainStd;zincBestTest;zincBestTrain
0;Utility regr.;;1;0.273;0.004;0.209;0.004;1;0;0.706;0.001;0.787;0.013;0;0;0.232;0.002;0.243;0.004;0;0;0.015;0.242;0.207;0.205;0;0;0.547;0.414;0.550;0.412;0;0
1;Rank regr.;;1;0.172;0.040;0.209;0.019;0;0;0.702;0.012;0.787;0.009;0;1;0.224;0.006;0.246;0.006;0;0;0.446;0.038;0.695;0.006;0;1;0.823;0.005;0.829;0.006;0;0
2;DirectRanker;;1;0.234;0.002;0.358;0.013;0;1;0.714;0.003;0.779;0.005;1;0;0.327;0.006;0.510;0.016;0;0;0.483;0.011;0.646;0.008;1;0;0.879;0.002;0.882;0.001;1;1
3;CmpNN;;1;0.195;0.007;0.352;0.004;0;0;0.632;0.076;0.704;0.061;0;0;0.381;0.008;0.563;0.024;1;1;0.351;0.055;0.568;0.030;0;0;0.819;0.002;0.823;0.002;0;0
4;Utility regr.;;1;0.469;0.056;0.456;0.081;0;0;0.729;0.006;0.809;0.015;1;0;0.353;0.052;0.451;0.091;0;0;0.243;0.125;0.748;0.069;0;0;0.790;0.003;0.787;0.003;0;0
5;Rank regr.;;1;0.481;0.014;0.621;0.002;0;1;0.717;0.011;0.845;0.006;0;1;0.310;0.017;0.423;0.021;0;0;0.495;0.021;0.823;0.064;0;1;0.827;0.011;0.828;0.010;0;0
6;DirectRanker;;1;0.502;0.028;0.589;0.011;0;0;0.712;0.007;0.799;0.004;0;0;0.429;0.026;0.545;0.023;0;0;0.439;0.065;0.749;0.056;0;0;0.894;0.012;0.903;0.015;1;1
7;CmpNN;;1;0.520;0.070;0.522;0.080;1;0;0.710;0.007;0.793;0.006;0;0;0.506;0.013;0.706;0.029;1;1;0.518;0.018;0.792;0.021;1;0;0.891;0.006;0.894;0.007;0;0
8;Utility regr.;;1;0.997;0.006;0.988;0.008;0;0;0.747;0.007;0.822;0.007;1;0;0.318;0.017;0.380;0.024;0;0;0.379;0.207;0.624;0.251;0;0;0.803;0.006;0.802;0.006;0;0
9;Rank regr.;;1;0.972;0.017;0.963;0.020;0;0;0.720;0.019;0.851;0.022;0;1;0.332;0.083;0.439;0.081;0;0;0.524;0.020;0.883;0.001;0;1;0.810;0.003;0.809;0.003;0;0
10;DirectRanker;;1;1.000;0.000;1.000;0.000;1;1;0.745;0.009;0.842;0.004;0;0;0.505;0.012;0.613;0.025;1;0;0.525;0.010;0.869;0.023;0;0;0.894;0.008;0.900;0.010;1;1
11;CmpNN;;1;1.000;0.000;1.000;0.000;1;1;0.718;0.020;0.770;0.038;0;0;0.503;0.010;0.666;0.013;0;1;0.527;0.064;0.849;0.038;1;0;0.873;0.002;0.876;0.002;0;0
\end{filecontents*}
\csvreader[
		column count=40,
		tabular={clrrrrr},
		separator=semicolon,
		table head={%
			& \multicolumn{1}{l}{} &%
			\multicolumn{1}{c}{\textbf{TRIANGLES}} &%
			\multicolumn{1}{c}{\textbf{OGB-molesol}} &%
			\multicolumn{1}{c}{\textbf{-mollipo}} &%
			\multicolumn{1}{c}{\textbf{-molfreesolv}} &%
			\multicolumn{1}{c}{\textbf{ZINC}}%
			\\\toprule%
		},
		before reading={\ifarxiv\setlength{\tabcolsep}{4pt}\fi},
		table foot=\bottomrule,
		late after line=\ifthenelse{\equal{\id}{4}\or\equal{\id}{8}}{\\\midrule}{\\},
		head to column names,
		filter=\equal{\isDefault}{1}
	]{data/results.csv}{}{%
		\ifthenelse{\equal{\id}{0}}{\multirow{4}{*}[0em]{\rotatebox[origin=c]{90}{\textsc{GCN}}\ifarxiv\else\hspace{0.5em}\fi}}{}%
		\ifthenelse{\equal{\id}{4}}{\multirow{4}{*}[0em]{\rotatebox[origin=c]{90}{\textsc{GIN}}\ifarxiv\else\hspace{0.5em}\fi}}{}%
		\ifthenelse{\equal{\id}{8}}{\multirow{4}{*}[0em]{\rotatebox[origin=c]{90}{\textsc{2-WL}}\ifarxiv\else\hspace{0.5em}\fi}}{}&%
		\textbf{\model} &%
		{\evalres{\triangleBestTest}{\triangleTestMean}{\triangleTestStd}} &%
		{\evalres{\molesolBestTest}{\molesolTestMean}{\molesolTestStd}} &%
		{\evalres{\mollipoBestTest}{\mollipoTestMean}{\mollipoTestStd}} &%
		{\evalres{\molfreesolvBestTest}{\molfreesolvTestMean}{\molfreesolvTestStd}} &%
		{\evalres{\zincBestTest}{\zincTestMean}{\zincTestStd}}%
	}}
\end{table*}

Looking at the results for the synthetic TRIANGLES dataset, we find that only the higher-order \ac{wl2gnn} is able to reliably rank graphs by their triangle counts.
This is plausible, because architectures bounded by the 1-\acs{wl} test, such as \ac{gcn} and \ac{gin}, are unable to detect cycles in graphs~\cite{Fuerer2017}.
While both the point- and the pair-wise \ac{wl2gnn} models achieve perfect or near-perfect $\tau_{B}$ scores on this task, the pair-wise approaches did perform more consistently, without a single inversion on the test graphs over 10 iterations of retraining.

Since the target graph rankings for all evaluated datasets are derived from regression values, all models have to learn a transitive preference relation.
Consequently, the ability of \ac{cmpnn}-based RankGNNs to learn non-transitive preferences is, in theory, not required to achieve optimal ranking performance.
If the sample size of training graph pairs is too small, such that it contains few transitivity-indicating subsets, e.g.\ $\{ G_1 \succeq G_2, G_2 \succeq G_3, G_1 \succeq G_3 \}$, the higher expressiveness of \acp{cmpnn} could even lead to overfitting and therefore worse generalization performance compared to DirectRanker.
Nonetheless, with the used sampling factor of $\alpha = 20$ (and $\alpha = 3$ for ZINC), each graph is, in expectation, sampled $40$ times ($6$ for ZINC).
This appears to be sufficient to prevent overfitting.
In fact, the \acs{cmpnn}-based RankGNNs perform very similarly to their DirectRanker-based counterparts.
However, since DirectRanker-based models allow for a more efficient sorting implementation than \acs{cmpnn}-based ones (cf.\ \cref{sec:rankgnn:sort}), we suggest the use of DirectRanker for problems where transitivity can be assumed. 

\subsection{Analysis of the Implicit Utilities of DirectRanker \acsp*{gnn}}%
\label{sec:eval:utilities}

\newcommand{\utilityPlot}[3][width=\utilityPlotSize]{\begin{tikzpicture}
    \ifarxiv
        \def\utilityPlotSize{0.4\linewidth}
    \else
        \def\utilityPlotSize{0.42\linewidth}
    \fi
	\begin{axis}[
		#1,
		height=\utilityPlotSize,
		enlargelimits=false,
		ymin=0, ymax=1,
		ticks=none,
		xlabel={target rank},
		ylabel={normalized utility},
		#2,
		label style={font=\scriptsize},
		title style={font=\small\bfseries}
	]
	    \addplot [mark=none, color=t_gray] table [x=i, y=point2, col sep=comma] {data/rank_utils_#3_ogb.csv};
		\addplot [mark=none, color=t_blue] table [x=i, y=point, col sep=comma] {data/rank_utils_#3_ogb.csv};
		\addplot [mark=none, color=t_red] table [x=i, y=pair, col sep=comma] {data/rank_utils_#3_ogb.csv};
		\addplot [line width=1pt, mark=none, color=black] table [x=i, y=target, col sep=comma] {data/rank_utils_#3_ogb.csv};
	\end{axis}
\end{tikzpicture}}
\begin{figure}[t]
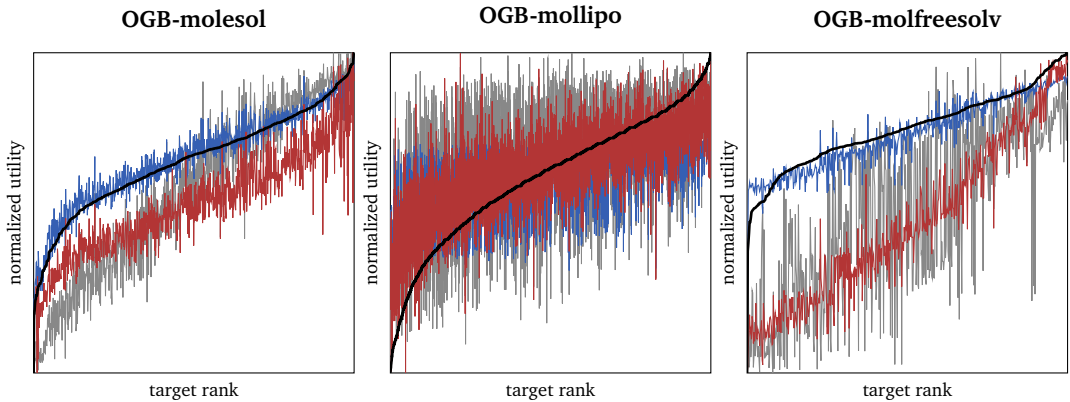

    \centering%
    \utilityPlot{title={OGB-molesol}}{wl2/ogbg-molesol}%
    \utilityPlot{title={OGB-mollipo}}{gin/ogbg-mollipo}%
    \utilityPlot{title={OGB-molfreesolv}}{wl2/ogbg-molfreesolv}%
    \caption{
        Normalized learned utility values of the point-wise \ac{gnn} regression model trained on the original utilities \textcolor{t_blue}{(in blue)}, the point-wise model trained on normalized ranks \textcolor{t_darkgray}{(in gray)} and the pair-wise DirectRanker model \textcolor{t_red}{(in red)}.
        For each dataset, we plot the predicted utilities of the \ac{gnn} architecture that achieved the best point-wise ranking performance in \cref{tbl:eval:results}, i.e.\ \ac{wl2gnn} for OGB-molesol and -molfreesolv and \ac{gin} for OGB-mollipo.
        Each point along the horizontal axes corresponds to a graph in the training split of a dataset.
        The graphs are sorted in ascending order by the ground truth utility values (shown in black) from which the target rankings are derived.
    }\label{fig:eval:utils}
\end{figure}
As described in \cref{sec:ranking:neural}, a DirectRanker model $f^{\text{DR}}_{\succeq}: O \times O \to {\{0,1\}}$ implicitly learns a utility function $f^{\text{DR}}_u: O \to \mathbb{R}$ from the set of pairs it sees during training.
We will now take a closer look at this implicitly learned utility function $f^{\text{DR}}_u$ and compare it to the explicitly learned utilities $f^{\text{util.}}_u$ and $f^{\text{rank}}_u$ of the point-wise \ac{gnn} regression models.
\Cref{fig:eval:utils} shows the values of all three, \textcolor{t_red}{$f^{\text{DR}}_u$ (in red)}, \textcolor{t_blue}{$f^{\text{util.}}_u$ (in blue)} and \textcolor{t_darkgray}{$f^{\text{rank}}_u$ (in gray)}, normalized to the unit interval.
Any monotonically increasing curve corresponds to a perfect ranking ($\tau_B = +1$), while a monotonically decreasing curve would signify an inverse ranking ($\tau_B = -1$).

As expected, the blue utility curves of the point-wise approaches align with the black target utility curves, while the gray curve more closely follows the 45° diagonal line on which the normalized graph ranks would lie.
However, this alignment does not necessarily imply good ranking performance.
For example, on the OGB-molfreesolv dataset, the \textcolor{t_blue}{blue utility curve} of the point-wise \ac{wl2gnn} model fits the black target curve fairly well for the graphs in the middle of the ranking.
However, near the low and the high graph ranks, the target curve abruptly falls/rises to its minimum and maximum values; the point-wise regression model that is trained on the original utilities ignores those outliers.
By instead training a point-wise model on the normalized ranks, outliers in the original utility values are effectively smoothed out, as can be seen in the \textcolor{t_darkgray}{gray} OGB-molfreesolv utility curve.
Looking at \cref{tbl:eval:results}, we find that this corresponds to a significantly higher mean $\tau_B$ coefficient and a lower variance on the OGB-molfreesolv dataset.
The pair-wise DirectRanker-based approach solves the problem of outliers in a more general fashion.
It uses a loss function that does not penalize for learning a monotonous, rank-preserving transformation of the target utility curve.
This allows it to effectively ``stretch'' the target utilities into a linearly growing curve with fewer abrupt changes, which results in a similar performance to that of the regression model trained on normalized ranks.

The target utilities of the OGB-molesol dataset are distributed more smoothly, without any outliers.
There the advantage of approaches that work well with outliers (e.g.\ pair-wise models) over the ones that do not is less pronounced.
Lastly, looking at the OGB-mollipo dataset, we also do not find outliers in the target utility curve.
However, there the pair-wise RankGNN models perform significantly better than the point-wise approaches.
The reason for this performance difference is not yet fully understood.

\section{Conclusion}%
\label{sec:conclusion}

In this paper, we addressed the problem of learning to rank graph-structured data and proposed RankGNNs, a combination of neural pair-wise ranking models and \acp{gnn}.
When compared with the na\"ive approach of using a point-wise \ac{gnn} regression model for ranking, we found that RankGNNs achieve a significantly higher or at least similar ranking performance on a variety of synthetic and real-world graph datasets.
We therefore conclude that RankGNNs are a promising approach for solving graph ranking problems.

There are various directions for future research.
First, due to the lack of graph ranking benchmark datasets, we had to use graph regression datasets in our evaluation instead.
For a more thorough analysis of the practical applicability of graph ranking models, a collection of real-world graph ranking benchmarks should be created.
One potential benchmark domain could, for example, be the drug screening problem we described in the introduction, where the training data consists of drug candidate pairs ranked by a human expert.

Second, list-wise graph ranking approaches could be evaluated in addition to the point- and pair-wise models considered in this paper.
Such list-wise models can be useful to learn a human's individual preferences for structured objects, such as task schedules or organizational hierarchies, represented as directed acyclic graphs or trees, respectively.
A list-wise ranking approach~\citep[e.g.][]{Pfannschmidt2018} would be able to consider context-dependent preferences in such scenarios~\citep{Huber1982}.
Yet another interesting idea, motivated by the behavior we observed for the point- and pair-wise \acs{wl2gnn}-based models on the OGB-molfreesolv dataset (cf.\ \Cref{fig:eval:utils}), is a hybrid approach that combines regression and ranking, that is, point-wise and pair-wise learning~\cite{sculley10}.

Third, although graph neural networks are quite popular these days, the problem of graph ranking could also be tackled by well-established kernel-based methods. In the past, there has been a lot of work on graph kernels \cite{vish_gk10}, making graph-structured data amenable to kernel-based learning methods. In principle, one may hence think of combining graph kernels with learning-to-rank methods such as RankSVM. However, our first experiences with an approach of that kind suggest that kernel-based approaches are computationally complex and do not scale sufficiently well, even for point-wise implementations\,---\,for larger data sets, the running time as well as the memory requirements are extremely high (which is also the reason why we excluded them from the experiments). Although they can be reduced using suitable approximation techniques, complexity clearly remains an issue. Besides, the ranking performance turned out to be rather poor. For pair-wise approaches, not only the complexity further increases, but the problem also becomes conceptually non-trivial. This is because the simple reduction of ranking to classification, on which RankSVM is based, no longer works (this reduction takes differences between feature vectors, an operation that cannot be applied to graphs). Instead, a (preference) kernel function on pairs of pairs of objects, i.e.\ on quadruples, has to be used \cite{waeg_kb09}. Nevertheless, this does of course not exclude the existence of more efficient (approximate) algorithms operating on kernel-representation for graphs. 
\long\def\acks#1{\section*{Acknowledgments}#1}
%\acks{This work was supported by German Research Foundation (DFG) within the Collaborative Research Center ``On-The-Fly Computing'' (SFB 901/3 project no.\ 160364472).}

% \vfill\pagebreak
% \appendix
% \input{content/appendix}
%\pagebreak
% \bibliographystyle{plainnat}
\bibliography{literature}

\begin{thebibliography}{29}
\providecommand{\natexlab}[1]{#1}
\providecommand{\url}[1]{\texttt{#1}}
\expandafter\ifx\csname urlstyle\endcsname\relax
  \providecommand{\doi}[1]{doi: #1}\else
  \providecommand{\doi}{doi: \begingroup \urlstyle{rm}\Url}\fi

\bibitem[Agarwal(2010)]{Agarwal2010}
Shivani Agarwal.
\newblock Learning to rank on graphs.
\newblock \emph{Machine Learning}, 81\penalty0 (3):\penalty0 333--357, 2010.
\newblock \doi{10.1007/s10994-010-5185-8}.

\bibitem[Zhang et~al.(2015)Zhang, Ji, Chen, Tang, Wang, Zhu, Jia, Cao, and
  Liu]{Zhang2015}
Wei Zhang, Lijuan Ji, Yanan Chen, Kailin Tang, Haiping Wang, Ruixin Zhu, Wei
  Jia, Zhiwei Cao, and Qi~Liu.
\newblock When drug discovery meets web search: Learning to rank for
  ligand-based virtual screening.
\newblock \emph{J. Cheminf.}, 7\penalty0 (1), 2015.

\bibitem[Joachims(2002)]{Joachims2002}
Thorsten Joachims.
\newblock Optimizing search engines using clickthrough data.
\newblock In \emph{Proceedings of the eighth {ACM} {SIGKDD} international
  conference on Knowledge discovery and data mining - {KDD}
  {\textquotesingle}02}. {ACM} Press, 2002.
\newblock \doi{10.1145/775047.775067}.

\bibitem[Burges et~al.(2005)Burges, Shaked, Renshaw, Lazier, Deeds, Hamilton,
  and Hullender]{Burges2005}
Chris Burges, Tal Shaked, Erin Renshaw, Ari Lazier, Matt Deeds, Nicole
  Hamilton, and Greg Hullender.
\newblock Learning to rank using gradient descent.
\newblock In \emph{I}, 2005.
\newblock \doi{10.1145/1102351.1102363}.

\bibitem[Burges(2010)]{Burges2010}
Chris Burges.
\newblock From {RankNet} to {LambdaRank} to {LambdaMART}: An overview.
\newblock Technical Report MSR-TR-2010-82, Microsoft Research, 2010.

\bibitem[K{\"{o}}ppel et~al.(2019)K{\"{o}}ppel, Segner, Wagener, Pensel,
  Karwath, and Kramer]{Koeppel2019}
Marius K{\"{o}}ppel, Alexander Segner, Martin Wagener, Lukas Pensel, Andreas
  Karwath, and Stefan Kramer.
\newblock Pairwise learning to rank by neural networks revisited:
  Reconstruction, theoretical analysis and practical performance.
\newblock In \emph{{ECML} {PKDD} 2019}, volume 11908 of \emph{Lecture Notes in
  Computer Science}, pages 237--252. Springer, 2019.

\bibitem[Rigutini et~al.(2011)Rigutini, Papini, Maggini, and
  Scarselli]{Rigutini2011}
L.~Rigutini, T.~Papini, M.~Maggini, and F.~Scarselli.
\newblock {SortNet}: Learning to rank by a neural preference function.
\newblock \emph{{IEEE} Transactions on Neural Networks}, 22\penalty0
  (9):\penalty0 1368--1380, 2011.
\newblock \doi{10.1109/tnn.2011.2160875}.

\bibitem[Mesaoudi{-}Paul et~al.(2018)Mesaoudi{-}Paul, H{\"{u}}llermeier, and
  Busa{-}Fekete]{mpub383}
A.~El Mesaoudi{-}Paul, E.\ H{\"{u}}llermeier, and R.\ Busa{-}Fekete.
\newblock Ranking distributions based on noisy sorting.
\newblock In \emph{Proc.\ ICML 2018, 35th International Conference on Machine
  Learning}, pages 3469--3477, Stockholm, Sweden, 2018.

\bibitem[Cao et~al.(2007)Cao, Qin, Liu, Tsai, and Li]{Cao2007}
Zhe Cao, Tao Qin, Tie-Yan Liu, Ming-Feng Tsai, and Hang Li.
\newblock Learning to rank.
\newblock In \emph{I}. {ACM} Press, 2007.
\newblock \doi{10.1145/1273496.1273513}.

\bibitem[Pfannschmidt et~al.(2018)Pfannschmidt, Gupta, and
  Hüllermeier]{Pfannschmidt2018}
Karlson Pfannschmidt, Pritha Gupta, and Eyke Hüllermeier.
\newblock Deep architectures for learning context-dependent ranking functions,
  March 2018.

\bibitem[Bruna et~al.(2013)Bruna, Zaremba, Szlam, and LeCun]{Bruna2013}
J.~Bruna, W.~Zaremba, A.~Szlam, and Y.~LeCun.
\newblock {Spectral Networks and Locally Connected Networks on Graphs}, 2013.

\bibitem[Henaff et~al.(2015)Henaff, Bruna, and LeCun]{Henaff2015}
Mikael Henaff, Joan Bruna, and Yann LeCun.
\newblock {Deep Convolutional Networks on Graph-Structured Data}, 2015.

\bibitem[Kipf and Welling(2017)]{Kipf2017}
T.~N. Kipf and M.~Welling.
\newblock Semi-supervised classification with graph convolutional networks.
\newblock \emph{ICLR}, 2017.

\bibitem[Zhang et~al.(2018)Zhang, Cui, Neumann, and Chen]{Zhang2018}
Muhan Zhang, Zhicheng Cui, Marion Neumann, and Yixin Chen.
\newblock An end-to-end deep learning architecture for graph classification.
\newblock In \emph{Thirty-Second AAAI Conference on Artificial Intelligence},
  2018.

\bibitem[Lee et~al.(2019)Lee, Lee, and Kang]{Lee2019}
Junhyun Lee, Inyeop Lee, and Jaewoo Kang.
\newblock Self-attention graph pooling.
\newblock In \emph{ICML}, pages 6661--6670, 2019.

\bibitem[Xu et~al.(2019)Xu, Hu, Leskovec, and Jegelka]{Xu2018}
K.~Xu, W.~Hu, J.~Leskovec, and S.~Jegelka.
\newblock {How Powerful are Graph Neural Networks?}
\newblock In \emph{ICLR}, 2019.

\bibitem[Cai et~al.(1992)Cai, Fürer, and Immerman]{Cai1992}
J.~Cai, M.~Fürer, and N.~Immerman.
\newblock An optimal lower bound on the number of variables for graph
  identification.
\newblock \emph{Combinatorica}, 12\penalty0 (4):\penalty0 389--410, 1992.

\bibitem[Damke et~al.(2020)Damke, Melnikov, and Hüllermeier]{Damke2020}
Clemens Damke, Vitalik Melnikov, and Eyke Hüllermeier.
\newblock A novel higher-order {Weisfeiler-Lehman} graph convolution.
\newblock In \emph{Proceedings of the 12th Asian Conference on Machine Learning
  (ACML 2020)}, volume 129 of \emph{Proceedings of Machine Learning Research}.
  PMLR, 2020.

\bibitem[Bouritsas et~al.(2020)Bouritsas, Frasca, Zafeiriou, and
  Bronstein]{Bouritsas2020}
Giorgos Bouritsas, Fabrizio Frasca, Stefanos Zafeiriou, and Michael~M.
  Bronstein.
\newblock Improving graph neural network expressivity via subgraph isomorphism
  counting, 2020.

\bibitem[Maron et~al.(2019)Maron, Ben{-}Hamu, Serviansky, and
  Lipman]{Maron2019}
Haggai Maron, Heli Ben{-}Hamu, Hadar Serviansky, and Yaron Lipman.
\newblock Provably powerful graph networks.
\newblock In \emph{NeurIPS 2019}, pages 2153--2164, 2019.

\bibitem[Hu et~al.(2020)Hu, Fey, Zitnik, Dong, Ren, et~al.]{Hu2020}
Weihua Hu, Matthias Fey, Marinka Zitnik, Yuxiao Dong, Hongyu Ren, et~al.
\newblock Open graph benchmark: Datasets for machine learning on graphs, 2020.

\bibitem[Sterling and Irwin(2015)]{Sterling2015}
Teague Sterling and John~J. Irwin.
\newblock {ZINC} 15 {\textendash} ligand discovery for everyone.
\newblock \emph{Journal of Chemical Information and Modeling}, 55\penalty0
  (11):\penalty0 2324--2337, 2015.
\newblock \doi{10.1021/acs.jcim.5b00559}.

\bibitem[Morris et~al.(2020)Morris, Kriege, Bause, Kersting, Mutzel, and
  Neumann]{Morris2020}
Christopher Morris, Nils~M. Kriege, Franka Bause, Kristian Kersting, Petra
  Mutzel, and Marion Neumann.
\newblock {TUDataset}: A collection of benchmark datasets for learning with
  graphs, 2020.

\bibitem[Kingma and Ba(2015)]{Kingma2015}
D.~P. Kingma and J.~Ba.
\newblock Adam: {A} method for stochastic optimization.
\newblock In \emph{ICLR}, 2015.

\bibitem[Fürer(2017)]{Fuerer2017}
M.~Fürer.
\newblock {On the Combinatorial Power of the Weisfeiler-Lehman Algorithm}.
\newblock In \emph{Lecture Notes in Computer Science}, pages 260--271. Springer
  International Publishing, 2017.

\bibitem[Huber et~al.(1982)Huber, Payne, and Puto]{Huber1982}
Joel Huber, John~W. Payne, and Christopher Puto.
\newblock Adding asymmetrically dominated alternatives: Violations of
  regularity and the similarity hypothesis.
\newblock \emph{Journal of Consumer Research}, 9\penalty0 (1):\penalty0 90,
  1982.
\newblock \doi{10.1086/208899}.

\bibitem[Sculley(2010)]{sculley10}
D.~Sculley.
\newblock Combined regression and ranking.
\newblock In \emph{Proceedings of the 16th {ACM} {SIGKDD} International
  Conference on Knowledge Discovery and Data Mining, Washington, DC, USA, July
  25-28, 2010}, pages 979--988, 2010.

\bibitem[Vishwanathan et~al.(2010)Vishwanathan, Schraudolph, Kondor, and
  Borgwardt]{vish_gk10}
S.V.N.\ Vishwanathan, N.M.\ Schraudolph, R.\ Kondor, and K.M.\ Borgwardt.
\newblock Graph kernels.
\newblock \emph{Journal of Machine Learning Research}, 11:\penalty0 1201--1242,
  2010.

\bibitem[Waegeman et~al.(2009)Waegeman, Baets, and Boullart]{waeg_kb09}
W.~Waegeman, B.~De Baets, and L.~Boullart.
\newblock Kernel-based learning methods for preference aggregation.
\newblock \emph{4OR}, 7:\penalty0 169--189, 2009.

\end{thebibliography}

\end{document}